\documentclass[sigconf,screen]{acmart}
\AtBeginDocument{%
  \providecommand\BibTeX{{%
    \normalfont B\kern-0.5em{\scshape i\kern-0.25em b}\kern-0.8em\TeX}}}

\copyrightyear{2024} 
\acmYear{2024} 
\setcopyright{rightsretained} 
\acmConference[Websci Companion '24]{16th ACM Web Science Conference}{May 21--24, 2024}{Stuttgart, Germany}
\acmBooktitle{16th ACM Web Science Conference (Websci Companion '24), May 21--24, 2024, Stuttgart, Germany}\acmDOI{10.1145/3630744.3663609}
\acmISBN{979-8-4007-0453-6/24/05}
\begin{document}

\title{Sexism Detection on a Data Diet}

\author{Rabiraj Bandyopadhyay}
\email{rabiraj.bandyopadhyay@gesis.org}
\affiliation{%
  \institution{GESIS Leibniz Institute for the Social Sciences}
  \streetaddress{Unter Sachssenhaussen 6-8}
  \city{Cologne}
  \country{Germany}
  \postcode{51105}
}

\author{Dennis Assenmacher}
\email{dennis.assenmacher@gesis.org}
\affiliation{%
  \institution{GESIS Leibniz Institute for the Social Sciences}
  \city{Cologne}
  \country{Germany}
}

\author{Jose M. Alonso-Moral}
\email{josemaria.alonso.moral@usc.es}
\affiliation{%
  \institution{Centro Singular de Investigaci\'{o}n en Tecnolox\'{i}as Intelixentes (CiTIUS), Universidade de Santiago de Compostela}
  \country{Spain}}

  \author{Claudia Wagner}
\email{claudia.wagner@gesis.org}
\affiliation{%
 \institution{GESIS Leibniz Institute for the Social Sciences, RWTH Aachen}
 \streetaddress{Cologne}
 \country{Germany}}

\renewcommand{\shortauthors}{Bandyopadhyay et al.}

\begin{abstract}
  There 
  is an increase in the proliferation of online hate 
  commensurate with the rise in the usage of social media. In response, there 
  is also a significant advancement in the creation of automated tools aimed at identifying harmful text content using approaches grounded in Natural Language Processing and Deep Learning. Although it is 
  known that training Deep Learning models require a substantial amount of annotated data, recent line of work suggests that models trained on specific subsets of the data still retain performance comparable to the model that was trained on the full dataset. 
  In this work, we  show how we can leverage influence scores to estimate the importance of a data point while training a model and designing a pruning strategy applied to the case of sexism detection. 
  We evaluate the model performance trained on data pruned with different pruning strategies on three out-of-domain datasets and find, that in accordance with other work a large fraction of instances can be removed without significant performance drop. However, we also discover that the strategies for pruning data, previously successful in Natural Language Inference tasks, do not readily apply to the detection of harmful content and instead amplify the already prevalent class imbalance even more, leading in the worst-case to a complete absence of the hateful class.\\
  
  \textcolor{red}{\textbf{Warning:} This paper contains instances of hateful
and sexist language to serve as examples.}
\end{abstract}






\maketitle

\section{Introduction}
Social media platforms, have evolved into vital instruments enabling individuals to maintain connections with others and express their views on various topics, including politics, technology, and everyday life. However, there has also been an increase in the amount of harmful content targeting people belonging to different demographics, race, religion, and sexual orientation; that is being generated by different users every day on these platforms (e.g., Twitter, Facebook, Reddit). As a natural consequence, it has become highly imperative to monitor (and potentially regulate) such harmful content. 

In this work we focus on sexism, which is widely defined as any
"prejudice, stereotyping, or discrimination, typically against women, on the basis of sex"\footnote{Oxford English Dictionary.}. Attempts have been made to detect sexism by using Natural Language Processing (NLP) techniques.
Curation of datasets \cite{waseem-hovy-2016-hateful, kirk-etal-2023-semeval, 10.1145/3599696.3612900} 
can also aid in sexism detection. With advancements made in Deep Learning (DL),
especially after the introduction of transformer architecture \cite{vaswani2017attention}, models like BERT \cite{DBLP:journals/corr/abs-1810-04805} or RoBERTa \cite{DBLP:journals/corr/abs-1907-11692} have become de-facto models that have been applied to detect sexism from text data \cite{sen-etal-2022-counterfactually, muti-etal-2023-uniboes, feely-etal-2023-qcon}. Even though the aforementioned 
publications use the whole dataset to train and evaluate their models, 
some researchers \cite{attendu-corbeil-2023-nlu, mishra-sachdeva-2020-need, azeemi-etal-2023-data, anand-etal-2023-influence, fayyaz2022bert} suggest that some data instances are more useful for driving the learning process and impacting the final model performance than others. Particularly, 
researchers in Computer Vision (CV) explore the usefulness of influence scores \cite{koh2020understanding, agarwal2022estimating, paul2021deep, pruthi2020estimating} to quantify the importance of a particular datapoint when gauging the performance of a model after training or fine-tuning (in case of pre-trained models). Influence scores use information present during the training process (e.g., confidence or gradient loss) to estimate the contribution of a datapoint to the final model performance.

We build upon the first insights from \citet{ethayarajh2022understanding}, \citet{fayyaz2022bert} and 
\citet{anand-etal-2023-influence} who first investigated influence scores and their usefulness for NLP problems. Our research specifically aims to examine the effects of various pruning strategies that utilize influence scores, with a particular focus on addressing sexism, a domain notably affected by significant class imbalance.



The rest of the manuscript is organized as follows.
Section \ref{sec:relw} provides readers with a brief review of the state of the art.
Section \ref{sec:infscores} introduces basic concepts to follow the rest of the work.
Section \ref{sec:methodology} presents material and methods to be used in the experiments.
Section \ref{sec:results} describes the main reported results.
Section \ref{sec:conc} summarizes concluding remarks and points out future work.
Finally, Section \ref{sec:limit} declares the limitations of this work.

\section{Related Work}
\label{sec:relw}
Training models in a data-efficient fashion and identifying important data points have always been one of the challenging problems in Machine Learning (ML) and attempts have been made to propose solutions to this problem leading to the rise of body of methods called robust statistics \cite{bookrobuststatistics}. 

\citet{koh2020understanding} extended robust statistics to study black-box predictions of Inception v3 network \cite{szegedyetal} on ImageNet dataset \cite{dengetal} using a formulation from robust statistics called influence functions. Other influence-based scores were also proposed, notable amongst them being forgetting scores \cite{toneva2019empirical}, GraNd and Error L2-Norm (EL2N) \cite{paul2021deep}, TracIn Scores \cite{pruthi2020estimating}, Variance of Gradients (VoG) \cite{agarwal2022estimating}, and Pointwise V-Information (PVI) \cite{ethayarajh2022understanding}. 

\citet{fayyaz2022bert} 
were the first who used influence scores in the NLP domain. They 
employed EL2N scores to identify the highest scoring data points within the SNLI and AGNews datasets. They found that models trained on approximately the top 70\% of the dataset, after pruning certain portions of the highest scoring examples from both datasets, achieved test accuracy scores surpassing those of models trained on the entirety of the dataset. \citet{attendu-corbeil-2023-nlu} designed a dynamic pruning strategy using a metric inspired from EL2N for binary classification and multi-class classification tasks on popular datasets like MNLI, SST-2, and QNLI.
They used their metric to retain a subset of the highest scoring examples after each epoch and found out that they 
could prune up to 50\% of the data and still retain performance as compared to the model trained on the full data. 

\citet{anand-etal-2023-influence} performed an extensive investigation of different influence scores on SNLI \cite{bowman2015large} and proprietary user-utterances datasets. They discovered that in context of both SNLI and user-utterances dataset, pruning the low scoring examples (which they termed easy examples) based on their VoG scores and training models on the resulting data led to an increase in test accuracy when compared to the random baseline. Pruning of hard examples led to a decrease in the model performance compared to random baseline. This confirmed the findings of \citet{sorscher2023neural} that hard data points contain critical information that can help a model make a decision regarding the decision boundary. However, for other influence scores that they tested on -- PVI \cite{ethayarajh2022understanding}, EL2N \cite{paul2021deep}, TracIn \cite{pruthi2020estimating} and Forget Scores \cite{toneva2019empirical}-- they found out that the pruning the harder or easier examples based on their respective scores did not lead to a gain in performance (measured by test accuracy) compared to random pruning but rather led to a sharp drop after a certain pruning rate (30\%).

In this study, we concentrate on three distinct influence scores that have shown the most promising results for Natural Language Understanding (NLU) tasks: PVI \cite{ethayarajh2022understanding}, EL2N \cite{paul2021deep}, and VoG \cite{agarwal2022estimating}. Each of these scores measures importance in a fundamentally different way, encompassing information-theoretic, margin-based, and gradient-based approaches, respectively.
To our best knowledge this is the first study that explores the utility of influence scores for the detection of hateful online communication. We refer the readers to Table \ref{tab:Table 10} for list of abbreviations and their corresponding full forms for convenience.

\section{Influence Scores}
\label{sec:infscores}

\subsection{Pointwise V-Information}
Pointwise V-Information (PVI)
is a method that extends the Pointwise Mutual Information (PMI) proposed in \cite{DBLP:journals/corr/abs-2002-10689} to NLP and tries to measure the usable bits of information for a model in predicting the corresponding label. The proposed method trains two models ($g$ and $g^{\prime}$), one on combination of  null inputs ($\phi$) and labels ($y$), and another on combination of text inputs ($x$) and labels ($y$) respectively to calculate the following quantities for each datapoint.
\begin{equation}
\begin{split}
&-log_{2}g[\phi][y] \\
&-log_{2}g^{\prime}[x][y]
\end{split}
\end{equation}

The PVI for a datapoint $x$ is 
calculated 
as follows:
\begin{equation}
   pvi(x) = -log_{2}g[\phi][y] + log_{2}g^{\prime}[x][y] 
\end{equation}

A negative PVI implies that the instance was harder for the model to predict \cite{ethayarajh2022understanding}. 
We closely follow the implementation as delineated in \cite{ethayarajh2022understanding} in calculating the PVI scores after training the model. More details about 
the experimental setup can be found in 
Section \ref{sec:methodology}. 

\subsection{Error L2-Norm}
Error L2-Norm (EL2N)
was introduced in \cite{paul2021deep}. As mentioned by \citet{paul2021deep, anand-etal-2023-influence,sorscher2023neural}, EL2N is a margin-based influence score, implying that the data points which are harder for the model to classify have high EL2N scores, and are closer to the decision boundary. Conversely 
data points that are easier for the model are farther away from the decision boundary. Hence EL2N scores
give us an idea about how easier or harder a particular data point was for a model. 
EL2N score was introduced as a metric to detect data points that can be pruned early in training \cite{paul2021deep} and was compared with forgetting scores \citet{toneva2019empirical}. \citet{anand-etal-2023-influence} found that EL2N is also an informative influence score that helps in pruning 
data points after training a model. The EL2N score for a data point $x$ is given by 
\begin{equation}
 \lVert softmax(g(x)) - y \rVert_{2}   
\end{equation}
where $g$ denotes the model. 

\subsection{Variance of Gradients}
Variance of Gradients (VoG)
introduced in \citet{agarwal2022estimating}, is another influence score that also assists in understanding the hardness or easiness of a datapoint in a training dataset. It was proposed as a ranking method to rank the hardness of examples for models trained on standardized CV datasets - CIFAR-10, CIFAR-100 \cite{krizhevskyetal2009}, and ImageNet \cite{dengetal2009}. 

VoG captures the \textit{"per example change in explanations over time"} \cite{agarwal2022estimating}, as opposed to saliency maps which scores the features of the input data based on their contribution to final output \cite{simonyan2014deep}. The method works by calculating the gradient of the pre-softmax activation with respect to the input at the true label position, then calculating the mean and variance of the gradient with respect to the input at each checkpoint and then taking an average over the per-pixel variance to get the final VoG of an input. \citet{agarwal2022estimating} also proposed VoG as a metric for performing data auditing and hence it aided them in identifying the noisy examples (corresponding to higher scores) from the aforementioned Image Recognition datasets, and they found out that removing the noisier examples led to a better performance of the models. 

In our case, since 
we work with language data, we calculated the gradient with respect to the input embeddings at each checkpoint, as done in \cite{anand-etal-2023-influence}. Mathematically, our method can be formulated as follows. 
Let $\mathbf{G}_{i}^{c}$ be the gradient with respect to input $i$ at checkpoint $c$. It is calculated with respect to pre-softmax output $\mathbf{A}_{i}^{c}$ with respect to the input embedding $\mathbf{E}_{i}^{c}$ using the following equation.
\begin{equation}
    \mathbf{G}_{i}^{c} = \frac{\partial{\mathbf{A}_{i}^{c}}}{\partial{\mathbf{E}_{i}^{c}}}
\end{equation} \\
The VoG score for each input example $i$ is computed as follows:
\begin{equation}
\begin{split}
\boldsymbol{\mu}_{i} &= \frac{1}{N_{c}}{\sum_{c} \mathbf{G}_{i}^{c}} \\
\mathbf{V}_{i} &= \frac{1}{\sqrt{N_{c}}} (\mathbf{G}_{i}^{c} - \boldsymbol{\mu}_{i})^{2}
\end{split}
\end{equation}
The unnormalized VoG score $v_{i}$ is calculated by taking the mean of $\mathbf{V}_{i}$ (average over the input embeddings \cite{anand-etal-2023-influence}).
For normalization of the VoG scores we use the class-normalized VoG prescribed by \cite{agarwal2022estimating}, which is calculated by computing the mean and standard deviation of VoG scores per class and then normalizing the VoG scores of each data point belonging to the corresponding class. Let the class mean be denoted as $\mu_{class}$ and standard deviation as $\sigma_{class}$. Then the normalized VoG for datapoint i belonging to that particular class is calculated as:
\begin{equation}
\begin{split}
    VoG_{i} = \frac{{v_{i}} - \mu_{class}} {\sigma_{class}}        
\end{split}
\end{equation}\\

It is worth noting that \citet{anand-etal-2023-influence} also introduced dataset normalized VoG scores on top of class-normalized VoG scores, but we experiment with the class normalized VoG score as prescribed in the original paper \cite{agarwal2022estimating}.\\

\section{Methodology}\label{sec:methodology}
\begin{figure*}[h!]
    \includegraphics[scale = 0.5]{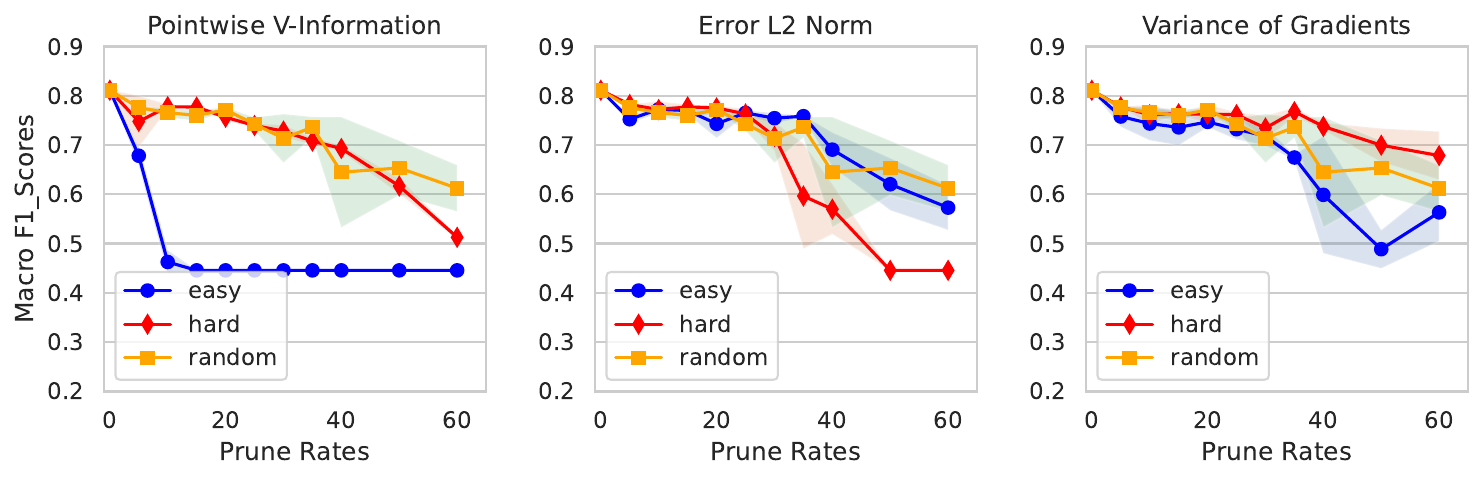}
    \caption{Macro F1-score on in-domain test data.}
    \label{fig:Figure 1}
\end{figure*}

\begin{figure*}[h!]
   \includegraphics[scale = 0.5]{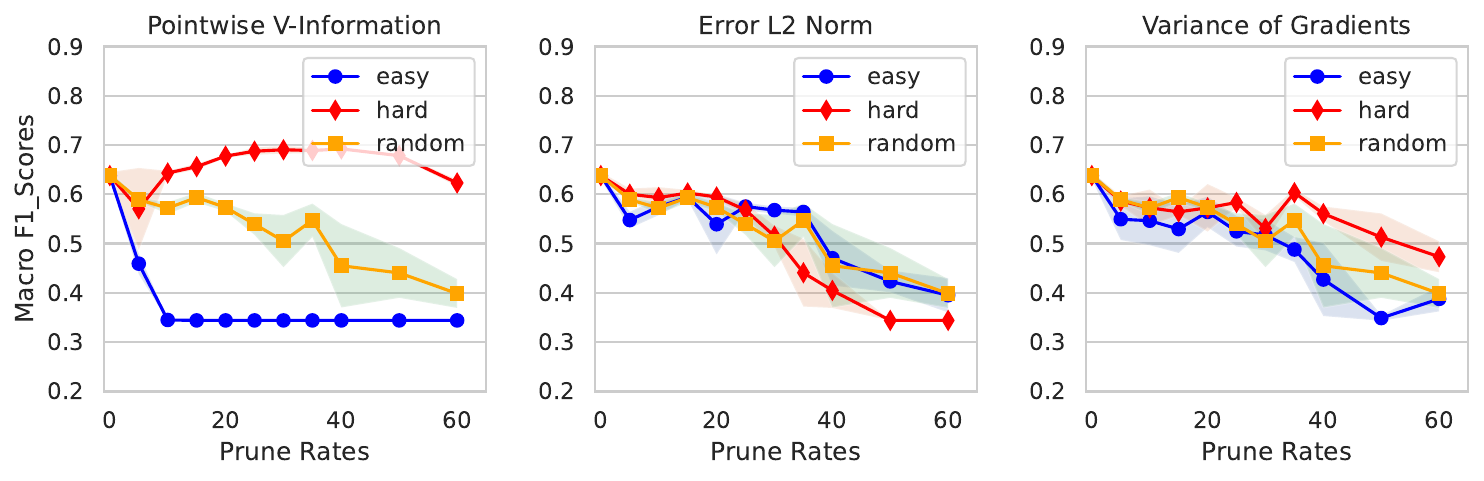}
    \caption{Macro F1-scores on EXIST dataset \cite{rodriguezsanchezetal}.}
    \label{fig:Figure 3}
\end{figure*}

\begin{figure*}[h!]
   \includegraphics[scale = 0.5]{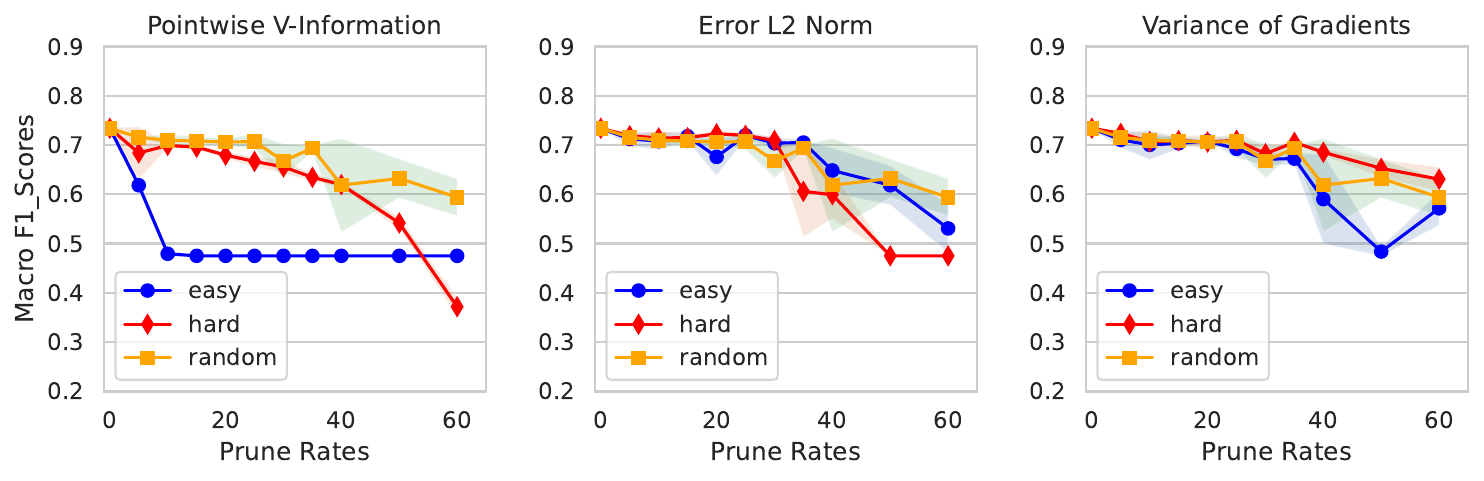}
    \caption{Macro F1-scores on Misogyny dataset \cite{guest-etal-2021-expert}.}
    \label{fig:Figure 4}
\end{figure*}

\subsection{Datasets}
For 
the experimental setup we created a training dataset by combining two well-known datasets -- EDOS (Explainable Detection of Online Sexism) dataset by \citet{kirk-etal-2023-semeval} and Call Me Sexist But dataset by \citet{samory2021call}. We will refer to the resulting dataset as Combined Data or in-domain data. We then split 
the resulting data into a 70/30 train/test split for in-domain evaluation. For cross-evaluation and determining the out-of-domain performance of 
various classifiers, we filtered data from the Hatecheck dataset \cite{rottger-etal-2021-hatecheck} based on target group \textbf{women}. Additionally we used the whole EXIST dataset \cite{rodriguezsanchezetal} and Misogyny dataset \cite{guest-etal-2021-expert} consisting of Reddit posts and comments to detect online misogyny. The class distribution of our training and out-of-domain evaluation datasets can be inspected in Table \ref{tab:table 3}.

\begin{table}[ht!]
\centering
    \begin{tabular}{lrr}
        \textbf{Dataset Name} & \textbf{Sexist} & \textbf{Non-Sexist} \\ [0.5ex]
        \hline
        {Combined Data (Train Split)} & {4676} & {18865}\\
        {Combined Data (Test Split)} & {1987} & {8103}\\
        {Hatecheck (Sexism) \cite{rottger-etal-2021-hatecheck}} & {373} & {136}\\
        {EXIST \cite{rodriguezsanchezetal}} & {1636} & {1800}\\
        {Misogyny \cite{guest-etal-2021-expert}} & {611} & {5777}\\
        \hline
    \end{tabular}
    \caption{Sexist and non-sexist counts in both in-domain and out-of-domain datasets.}
    \label{tab:table 3}
\end{table}

\subsection{Model and Settings}
For testing the influence scores based pruning we used the BERT \cite{devlin-etal-2019-bert} model variant \textit{bert-base-cased} from Huggingface\footnote{\url{https://huggingface.co/docs/transformers/model_doc/bert}} and AdamW Optimizer \cite{loshchilov2019decoupled} with the parameter settings summarized in Table \ref{tab:Table 1}.

\begin{table}[h!]
    \centering
    \begin{tabular}{lc}
        \hline
        \textbf{Hyperparameter} & \textbf{Value}\\ [0.5ex] 

        Learning Rate & 2e-6 \\ 
  
        Epochs & 10 \\
     
        Scheduler & Linear\\
    
        Batch size & 32\\
        \hline
    \end{tabular}
    \caption{Hyperparameter settings.}
    \label{tab:Table 1}
\end{table}

We saved the weights of the model (checkpoint) after every epoch and calculated the PVI \cite{ethayarajh2022understanding} and EL2N \cite{paul2021deep} scores as these 
two scores assist us in understanding the behaviour of data points at each epoch of a model training \cite{anand-etal-2023-influence, paul2021deep, toneva2019empirical}. VoG scores \cite{agarwal2022estimating} were calculated after the end of the training. We performed this step 
three times and took an average of the influence scores for each datapoint.
After the scores were calculated, we sorted scores from the hardest to the easiest (i.e., from the lowest scoring to the highest scoring in case of PVI, and from the highest scoring to the lowest scoring in case of EL2N and VoG respectively). We used the following pruning rates -- $(5\%, 10 \%, 15 \%, 20 \%, 25 \%, 30 \%, 35\%,40\%,50\%, 60\%)$ -- to prune data for 
the experiments after sorting the data from the hardest to the easiest for all the influence scores. In the case of pruning hard examples, we pruned from the top for all the influence scores and from the bottom in case of 
easy examples.  We fine-tuned pre-trained BERT models \cite{devlin-etal-2019-bert} on each of the resultant pruned datasets (easy and hard), with the 
hyperparameter setting 
in Table \ref{tab:Table 2}.

\begin{table}[h!]
    \centering
    \begin{tabular}{lc}
        \hline
        \textbf{Hyperparameter} & \textbf{Value}\\ [0.5ex] 
        Learning Rate & 1e-6 \\ 
        Epochs & 5 \\
        Scheduler & Linear\\
        Batch size & 32\\
        \hline      
    \end{tabular}
    \caption{Hyperparameter settings (training on pruned data).}
    \label{tab:Table 2}
\end{table}

Furthermore, we randomly pruned the training data with the pruning rates as mentioned before and fine-tuned pre-trained BERT models on them to compare performance. We chose the hyperparameter settings as shown in Table \ref{tab:Table 2} because through empirical observation we found out that the models 
tend to overfit on the pruned data, so we decided to lower the learning rate and train it for lesser epochs. All the models were trained on a single 40 GB partition of an NVIDIA A100 GPU.

\begin{figure*}[h!]
    \includegraphics[scale = 0.5]{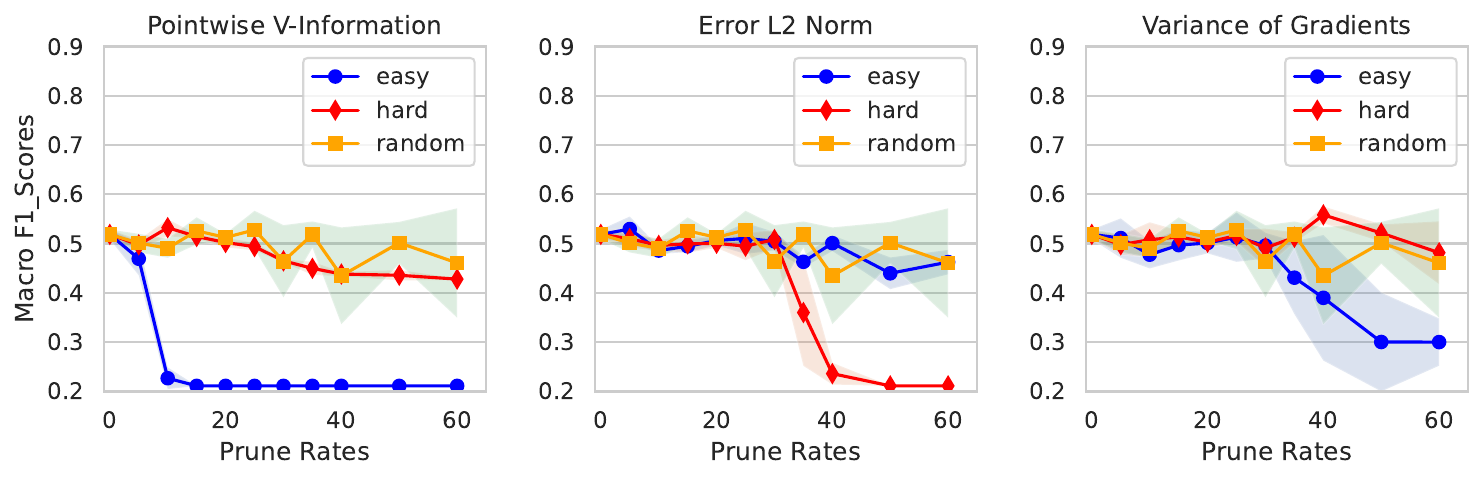}
    \caption{Macro F1-scores on Hatecheck instances containing the identity-term "woman" \cite{rottger-etal-2021-hatecheck}.}
    \label{fig:Figure 2}
\end{figure*}

\begin{figure*}[h!]
    \includegraphics[scale = 0.5]{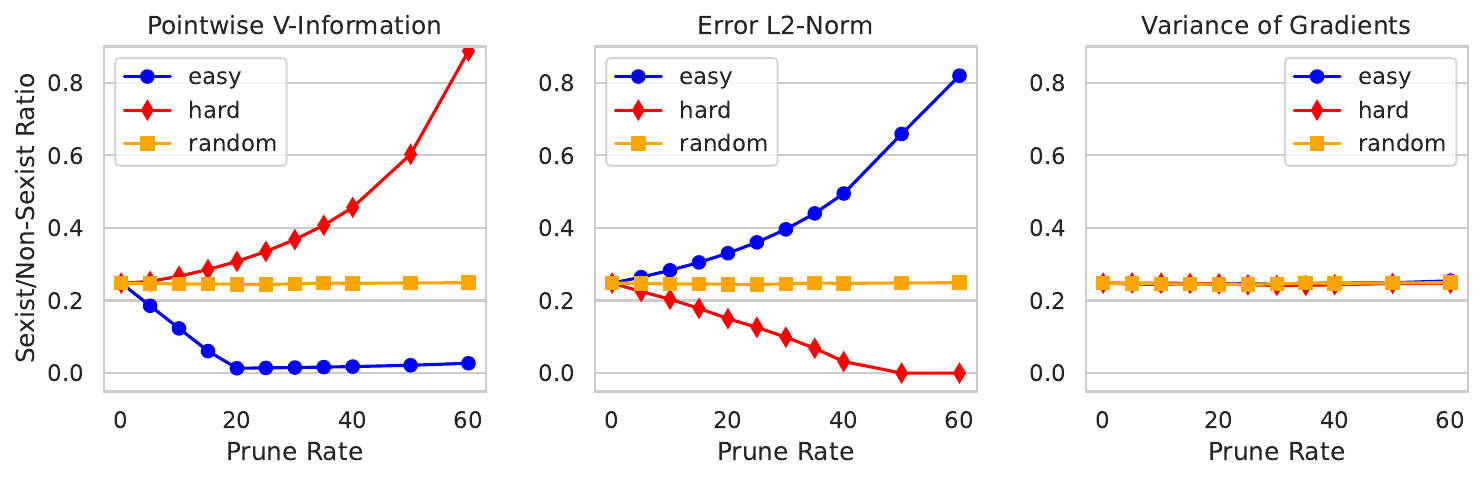}
    \caption{Sexist to non-sexist data ratio for each of the pruning rates and influence scores.}
    \label{fig:Figure 5}
\end{figure*}

\section{Results}\label{sec:results}


\subsection{General Findings}\label{subsec: general_findings}

Figure \ref{fig:Figure 1} shows the macro-F1 scores of models trained on datasets at different pruning rates and different pruning types (only easy instances, only hard instances, and random). One of our key observations is that randomly removing up to 50\% of instances from 
the training data does not significantly affect performance. In other words, we can eliminate a large portion of 
data instances without impacting the results. This is in line with \citet{anand-etal-2023-influence} who found similar patterns for the SNLI dataset. Additionally, we observed that pruning the easy examples and training the models on the remaining examples did not lead in general to improvements over random pruning. We observed a similar trend for the out-of-domain performance on both the EXIST and Misogyny dataset, evident in Figures \ref{fig:Figure 3} and \ref{fig:Figure 4}, respectively. 

Furthermore, for the Hatecheck test data, we discovered that the models generally do not achieve a satisfying out-of-domain performance even when trained on the combined data. This may be caused due to the fact that the evaluation data is particularly hard as all of its instances contain the identity term \textit{woman}. We know from previous studies that models predicting sexism are prone to learn spurious correlations (associating the term 
\textit{woman} with the positive class). Hence, they fail in this particular setup. Moreover, Hatecheck was 
developed as a stress test for models, already consisting of more complicated examples (e.g., containing leet speech or misspellings).
Additionally, the pruning of the easy and hard data points based on any of the influence scores did not marginally improve upon the random pruning baseline performance, as evident in Figure \ref{fig:Figure 2}. 

\subsection{Particular Findings from Influence Scores}\label{subsec: inf_scores_results}

Here, will discuss particular observations for each influence score individually.
 
\textbf{PVI:} When we increased the pruning of easy examples, we consistently observed a significant drop in performance for both in-domain and out-of-domain test sets.

On closer inspection, we discovered with increased pruning rates of easy examples more class-imbalance was introduced to the training data (see Figure \ref{fig:Figure 5}). By applying a pruning rate of 20\%, we effectively eliminated almost all sexist instances from the dataset, resulting in the model receiving insufficient data from the positive class to establish a clear decision boundary. When we pruned the hard examples, we observed that it did not exacerbate the imbalance problem. For the in-domain performance, we observed that pruning the difficult data points did not lead to performance gain over random pruning. A similar pattern was observed for the Misogyny dataset: pruning difficult examples did not enhance performance compared to random pruning. In fact, performance declined with the removal of more data, as illustrated in Figure \ref{fig:Figure 4}. However, we observed that pruning the difficult data points and training the models on the pruned data resulted in performance gain over random pruning in the case of EXIST dataset \cite{rodriguezsanchezetal}; this trend is depicted in Figure \ref{fig:Figure 3}.\\

\textbf{EL2N:} As we pruned the hard examples based on EL2N scores from the training data, we discovered that at pruning rates 
from 5\% to 25\% the performance closely follows that of random pruning in both in-domain and out-of-domain test sets.
With pruning rates greater than 25\% we effectively eliminated most of the sexist examples as evident from Figure \ref{fig:Figure 5} leading to an absence of examples of the positive class that can assist the models in establishing a decision boundary resulting in a drop in performance.

When we pruned the easy examples, we observed that it did not intensify the imbalance problem. In the case of in-domain performance, we observed that pruning the easy data points did not lead to a gain in performance over random baseline. A similar trend was observed regarding performance in all 
the out-of-domain datasets as illustrated in Figures  \ref{fig:Figure 3}, \ref{fig:Figure 4}, and \ref{fig:Figure 2}.


\textbf{VoG:} On pruning the hard data points based on VoG scores, we did not observe any performance improvement over random pruning in 
rates 
from 5\% to 30\% for in-domain as well as out-of-domain test sets. In the case of in-domain data, on further pruning the hard data points we did observe an improvement in performance but with a high margin of error when compared to random pruning. A similar trend was observed in out-of-domain performance as well. On pruning the easy examples, we did not observe any improvement in performance compared to random pruning baseline for the pruning rates 
from 5\% to 30\%. 

Further pruning of the easy examples led to a significant drop in performance when compared to the random pruning baseline in case of in-domain performance. We also observed similar trends in terms of performance in all the out-of-domain test sets. 
These findings were in contrast to findings of \cite{anand-etal-2023-influence} who found out that pruning easy examples from SNLI based on their VoG scores led to better performance when compared to random pruning in terms of test accuracy.

\subsection{Analysis of examples after fine-tuning}\label{subsec: analysis}

\begin{figure*}[h!]
    \includegraphics[scale = 0.5]{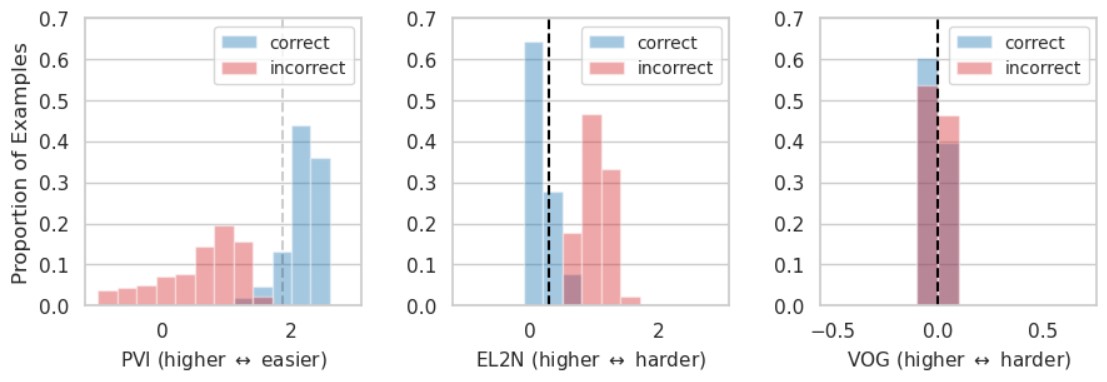}
    \caption{\small{Proportion of correct and incorrectly classified sexist examples after training for each influence score.}}
    \label{fig:Figure 6}
\end{figure*}

In this section we 
provide the distribution of the 3 influence scores for the sexist datapoints that were calculated after fine-tuning of the BERT model \cite{devlin-etal-2019-bert} on our training data. Figure \ref{fig:Figure 6} showcases the distribution of the misclassified and classified sexist data points in the train split of the combined data after training the models. We observed that EL2N and VOG 
have a distinct boundary for classified and misclassified data points. VoG scores on the other hand resulted in no distinct boundary.

\begin{table}[h!]
    \footnotesize
    \begin{tabular}{p{6cm} p{0.7cm} p{0.8cm}}
         \textbf{text} & \textbf{gold label} & \textbf{score}\\
         \hline
         MENTION1047 MENTION1497 pls don't the Trump womrn supporters paid or no r so much hotter <hyperlink>
         & sexist & -6.19671\\
         \hline
         SESSIONS IS A FUCKIN PUSSY!! CASE CLOSED!! END OF STORY!! & sexist & -5.80508\\
         \hline
         lmfao i didnt even notice that. I saw she/ and my brain immediately classified that person as retarded and ignored the rest... & sexist & -5.61337\\
         \hline
         I wonder which of the blonde ones screwed Colin to get that score? Probably both. & sexist & -5.16273 \\
         \hline
         Here are the reasons men get married: She threatens to cut off the pussy. Your family says you ain't a real man. You want to make her haaappy. & non-sexist & -5.15521\\
         \hline
    \end{tabular}
    \caption{\small{Top 5 low scoring examples (hard) based on PVI scores.}}
    \label{tab:table 4}
\end{table}

We also present the top-5 high scoring and low scoring examples which were obtained after calculating the PVI scores in Tables \ref{tab:table 4} and \ref{tab:table 5}. Example instances for the remaining metrics can be
inspected in Appendix \ref{app: App A.1}. We chose to elaborate on the findings of the PVI scores, as we observe the most interesting deviations from random pruning, as described in the previous section. 
Several noteworthy characteristics emerge in the more challenging examples listed in Table \ref{tab:table 4}. For instance, the data point with the lowest score includes a misspelling of "woman" introducing noise into the text that complicates model interpretation. Furthermore, the examples ranked $3^{rd}$ and $4^{th}$ exhibit implicit sexism, posing classification challenges for the model as well.
The last example, marked as non-sexist, posed a challenge for our models due to the presence of a trigger word ('pussy'). This instance could also be seen as a borderline case, potentially warranting a sexist classification.

Looking at the examples in Table \ref{tab:table 5} we observed that they were relatively easy for the model to classify because of presence of dedicated trigger word (explicit sexism). We further investigated the misclassified sexist instances of the training dataset with negative scores and found out that majority were from EDOS dataset. The same phenomenon was observed for misclassified non-sexist instances. 

\begin{table}[h!]
    \footnotesize
    \begin{tabular}{p{6cm} p{0.7cm} p{0.8cm}}
         \textbf{text} & \textbf{gold label} & \textbf{score}\\
         \hline
         I said in my family dude not every fucking black guy in the world <emojis>. If they blame women they are weak bitches.
         & sexist & 2.35117\\
         \hline
         If a bitch hit me like that she better knuckle the fuck up. & sexist & 2.35122\\
         \hline
         On top of everything else this bitch is down right ugly. Probably why she hates white guys because she is not attractive to us. & sexist & 2.35144\\
         \hline
         No, there are none. Period. LTR to your heart's content but do not wife a bitch & sexist & 2.35144 \\
         \hline
         Yeah dude, most blondes I know put out more and prefer men of color. In is so about the outer game. That being said most blondes are more likely to be bitches if you don't come at them right. & sexist & 2.35166\\
         \hline
    \end{tabular}
    \caption{\small{Top 5 high scoring examples (easy) based on PVI scores.}}
    \label{tab:table 5}
\end{table}




\section{Conclusions and future work}
\label{sec:conc}
In this work, we have used several influence-functions in the domain of sexism detection and evaluated their usefulness in the context of data pruning. As a main result, we found that we can remove up to 50\% of the data without real impact on the out-of-domain performance on all datasets, indicating that the training data might contain a lot of information that either confuses the model or information that is not relevant for the task \cite{anand-etal-2023-influence, fayyaz2022bert, attendu-corbeil-2023-nlu}. 

In contrast to other work, we do not recommend 
relying on pruning strategies that focus on the removal of only easy or hard instances, as this may lead to catastrophic class-imbalance. Rather, we need to design strategies to sample data from datasets based on their influence scores, especially when dealing with noisy and imbalanced datasets that are used to study hate speech and related constructs \cite{madukwe-etal-2020-data}. Furthermore, we observed that most of the sexist instances in our training data are considered PVI-easy or EL2N-hard. If EL2N-hard-instances are really \textit{"examples that contain critical information about the decision boundaries of classes in larger, less noisy datasets"} \cite{anand-etal-2023-influence}, we either have training data with significant noise (misannotations) or the datasets in general lack of high-quality examples that help to shape the decision boundaries. 

We hope that our experiments help in steering the field of hate speech detection towards the inclusion of influence scores in determining the quality of datasets that are usually curated for hate speech detection. Future work will incorporate what we learned from this study and design better sampling strategies that do not worsen the imbalance problem. In the future, we will also look at designing pruning strategies that can utilize the dynamic nature of the influence scores (especially PVI and EL2N) in informing the selection of data points. On further investigation we found out that most of the easy instances were all from the EDOS dataset, whereas the hard examples contained examples from both EDOS and Call Me Sexist But dataset.

\section{Limitations}
\label{sec:limit}

This study is not without limitations. First, merging two sexist datasets for the creation of our training set may have led to a reduction of the general quality due to different construct operationalizations.  Additionally, we did not experiment with different hyperparameter optimization techniques while training the models on the dataset and went forward with default settings. We note that the main purpose of this work was not to show the best performing influence score; rather this work serves as a proof-of-concept in using diverse set of influence scores in scoring the effect of data points after training or fine-tuning a model.   

\begin{acks}
The authors would take this opportunity to thank the funding from the Horizon Europe research and innovation program under the Marie
Skłodowska-Curie Grant Agreement No. 101073351 and the department of Computational Social Science at GESIS for providing the infrastructure to run the experiments.
This work is also partially supported by project DeepR3 (Grant TED2021-130295B-C33 funded by MCIN/AEI/10.13039/501100011033 and by the “European Union NextGenerationEU/PRTR”).  
\end{acks}

\bibliographystyle{ACM-Reference-Format}
\bibliography{citations}


\begin{thebibliography}{36}


\ifx \showCODEN    \undefined \def \showCODEN     #1{\unskip}     \fi
\ifx \showDOI      \undefined \def \showDOI       #1{#1}\fi
\ifx \showISBNx    \undefined \def \showISBNx     #1{\unskip}     \fi
\ifx \showISBNxiii \undefined \def \showISBNxiii  #1{\unskip}     \fi
\ifx \showISSN     \undefined \def \showISSN      #1{\unskip}     \fi
\ifx \showLCCN     \undefined \def \showLCCN      #1{\unskip}     \fi
\ifx \shownote     \undefined \def \shownote      #1{#1}          \fi
\ifx \showarticletitle \undefined \def \showarticletitle #1{#1}   \fi
\ifx \showURL      \undefined \def \showURL       {\relax}        \fi
\providecommand\bibfield[2]{#2}
\providecommand\bibinfo[2]{#2}
\providecommand\natexlab[1]{#1}
\providecommand\showeprint[2][]{arXiv:#2}

\bibitem[Agarwal et~al\mbox{.}(2022)]%
        {agarwal2022estimating}
\bibfield{author}{\bibinfo{person}{Chirag Agarwal}, \bibinfo{person}{Daniel D'souza}, {and} \bibinfo{person}{Sara Hooker}.} \bibinfo{year}{2022}\natexlab{}.
\newblock \bibinfo{title}{Estimating Example Difficulty Using Variance of Gradients}.
\newblock
\newblock
\showeprint[arxiv]{2008.11600}~[cs.CV]


\bibitem[Anand et~al\mbox{.}(2023)]%
        {anand-etal-2023-influence}
\bibfield{author}{\bibinfo{person}{Nikhil Anand}, \bibinfo{person}{Joshua Tan}, {and} \bibinfo{person}{Maria Minakova}.} \bibinfo{year}{2023}\natexlab{}.
\newblock \showarticletitle{Influence Scores at Scale for Efficient Language Data Sampling}. In \bibinfo{booktitle}{\emph{Proceedings of the 2023 Conference on Empirical Methods in Natural Language Processing}}, \bibfield{editor}{\bibinfo{person}{Houda Bouamor}, \bibinfo{person}{Juan Pino}, {and} \bibinfo{person}{Kalika Bali}} (Eds.). \bibinfo{publisher}{Association for Computational Linguistics}, \bibinfo{address}{Singapore}, \bibinfo{pages}{2485--2510}.
\newblock
\urldef\tempurl%
\url{https://doi.org/10.18653/v1/2023.emnlp-main.152}
\showDOI{\tempurl}


\bibitem[Attendu and Corbeil(2023)]%
        {attendu-corbeil-2023-nlu}
\bibfield{author}{\bibinfo{person}{Jean-michel Attendu} {and} \bibinfo{person}{Jean-philippe Corbeil}.} \bibinfo{year}{2023}\natexlab{}.
\newblock \showarticletitle{{NLU} on Data Diets: Dynamic Data Subset Selection for {NLP} Classification Tasks}. In \bibinfo{booktitle}{\emph{Proceedings of The Fourth Workshop on Simple and Efficient Natural Language Processing (SustaiNLP)}}, \bibfield{editor}{\bibinfo{person}{Nafise Sadat~Moosavi}, \bibinfo{person}{Iryna Gurevych}, \bibinfo{person}{Yufang Hou}, \bibinfo{person}{Gyuwan Kim}, \bibinfo{person}{Young~Jin Kim}, \bibinfo{person}{Tal Schuster}, {and} \bibinfo{person}{Ameeta Agrawal}} (Eds.). \bibinfo{publisher}{Association for Computational Linguistics}, \bibinfo{address}{Toronto, Canada (Hybrid)}, \bibinfo{pages}{129--146}.
\newblock
\urldef\tempurl%
\url{https://doi.org/10.18653/v1/2023.sustainlp-1.9}
\showDOI{\tempurl}


\bibitem[Azeemi et~al\mbox{.}(2023)]%
        {azeemi-etal-2023-data}
\bibfield{author}{\bibinfo{person}{Abdul Azeemi}, \bibinfo{person}{Ihsan Qazi}, {and} \bibinfo{person}{Agha Raza}.} \bibinfo{year}{2023}\natexlab{}.
\newblock \showarticletitle{Data Pruning for Efficient Model Pruning in Neural Machine Translation}. In \bibinfo{booktitle}{\emph{Findings of the Association for Computational Linguistics: EMNLP 2023}}, \bibfield{editor}{\bibinfo{person}{Houda Bouamor}, \bibinfo{person}{Juan Pino}, {and} \bibinfo{person}{Kalika Bali}} (Eds.). \bibinfo{publisher}{Association for Computational Linguistics}, \bibinfo{address}{Singapore}, \bibinfo{pages}{236--246}.
\newblock
\urldef\tempurl%
\url{https://doi.org/10.18653/v1/2023.findings-emnlp.18}
\showDOI{\tempurl}


\bibitem[Bertaglia et~al\mbox{.}(2023)]%
        {10.1145/3599696.3612900}
\bibfield{author}{\bibinfo{person}{Thales Bertaglia}, \bibinfo{person}{Katarina Bartekova}, \bibinfo{person}{Rinske Jongma}, \bibinfo{person}{Stephen Mccarthy}, {and} \bibinfo{person}{Adriana Iamnitchi}.} \bibinfo{year}{2023}\natexlab{}.
\newblock \showarticletitle{Sexism in Focus: An Annotated Dataset of YouTube Comments for Gender Bias Research}. In \bibinfo{booktitle}{\emph{Proceedings of the 3rd International Workshop on Open Challenges in Online Social Networks}} (Rome, Italy) \emph{(\bibinfo{series}{OASIS '23})}. \bibinfo{publisher}{Association for Computing Machinery}, \bibinfo{address}{New York, NY, USA}, \bibinfo{pages}{22–28}.
\newblock
\showISBNx{9798400702259}
\urldef\tempurl%
\url{https://doi.org/10.1145/3599696.3612900}
\showDOI{\tempurl}


\bibitem[Bowman et~al\mbox{.}(2015)]%
        {bowman2015large}
\bibfield{author}{\bibinfo{person}{Samuel~R. Bowman}, \bibinfo{person}{Gabor Angeli}, \bibinfo{person}{Christopher Potts}, {and} \bibinfo{person}{Christopher~D. Manning}.} \bibinfo{year}{2015}\natexlab{}.
\newblock \bibinfo{title}{A large annotated corpus for learning natural language inference}.
\newblock
\newblock
\showeprint[arxiv]{1508.05326}~[cs.CL]


\bibitem[Deng et~al\mbox{.}(2009a)]%
        {dengetal}
\bibfield{author}{\bibinfo{person}{Jia Deng}, \bibinfo{person}{Wei Dong}, \bibinfo{person}{Richard Socher}, \bibinfo{person}{Li-Jia Li}, \bibinfo{person}{Kai Li}, {and} \bibinfo{person}{Li Fei-Fei}.} \bibinfo{year}{2009}\natexlab{a}.
\newblock \showarticletitle{ImageNet: A large-scale hierarchical image database}. In \bibinfo{booktitle}{\emph{2009 IEEE Conference on Computer Vision and Pattern Recognition}}. \bibinfo{pages}{248--255}.
\newblock
\urldef\tempurl%
\url{https://doi.org/10.1109/CVPR.2009.5206848}
\showDOI{\tempurl}


\bibitem[Deng et~al\mbox{.}(2009b)]%
        {dengetal2009}
\bibfield{author}{\bibinfo{person}{Jia Deng}, \bibinfo{person}{Wei Dong}, \bibinfo{person}{Richard Socher}, \bibinfo{person}{Li-Jia Li}, \bibinfo{person}{Kai Li}, {and} \bibinfo{person}{Li Fei-Fei}.} \bibinfo{year}{2009}\natexlab{b}.
\newblock \showarticletitle{ImageNet: A large-scale hierarchical image database}. In \bibinfo{booktitle}{\emph{2009 IEEE Conference on Computer Vision and Pattern Recognition}}. \bibinfo{pages}{248--255}.
\newblock
\urldef\tempurl%
\url{https://doi.org/10.1109/CVPR.2009.5206848}
\showDOI{\tempurl}


\bibitem[Devlin et~al\mbox{.}(2018)]%
        {DBLP:journals/corr/abs-1810-04805}
\bibfield{author}{\bibinfo{person}{Jacob Devlin}, \bibinfo{person}{Ming{-}Wei Chang}, \bibinfo{person}{Kenton Lee}, {and} \bibinfo{person}{Kristina Toutanova}.} \bibinfo{year}{2018}\natexlab{}.
\newblock \showarticletitle{{BERT:} Pre-training of Deep Bidirectional Transformers for Language Understanding}.
\newblock \bibinfo{journal}{\emph{CoRR}}  \bibinfo{volume}{abs/1810.04805} (\bibinfo{year}{2018}).
\newblock
\showeprint[arXiv]{1810.04805}
\urldef\tempurl%
\url{http://arxiv.org/abs/1810.04805}
\showURL{%
\tempurl}


\bibitem[Devlin et~al\mbox{.}(2019)]%
        {devlin-etal-2019-bert}
\bibfield{author}{\bibinfo{person}{Jacob Devlin}, \bibinfo{person}{Ming-Wei Chang}, \bibinfo{person}{Kenton Lee}, {and} \bibinfo{person}{Kristina Toutanova}.} \bibinfo{year}{2019}\natexlab{}.
\newblock \showarticletitle{{BERT}: Pre-training of Deep Bidirectional Transformers for Language Understanding}. In \bibinfo{booktitle}{\emph{Proceedings of the 2019 Conference of the North {A}merican Chapter of the Association for Computational Linguistics: Human Language Technologies, Volume 1 (Long and Short Papers)}}, \bibfield{editor}{\bibinfo{person}{Jill Burstein}, \bibinfo{person}{Christy Doran}, {and} \bibinfo{person}{Thamar Solorio}} (Eds.). \bibinfo{publisher}{Association for Computational Linguistics}, \bibinfo{address}{Minneapolis, Minnesota}, \bibinfo{pages}{4171--4186}.
\newblock
\urldef\tempurl%
\url{https://doi.org/10.18653/v1/N19-1423}
\showDOI{\tempurl}


\bibitem[Ethayarajh et~al\mbox{.}(2022)]%
        {ethayarajh2022understanding}
\bibfield{author}{\bibinfo{person}{Kawin Ethayarajh}, \bibinfo{person}{Yejin Choi}, {and} \bibinfo{person}{Swabha Swayamdipta}.} \bibinfo{year}{2022}\natexlab{}.
\newblock \bibinfo{title}{Understanding Dataset Difficulty with $\mathcal{V}$-Usable Information}.
\newblock
\newblock
\showeprint[arxiv]{2110.08420}~[cs.CL]


\bibitem[Fayyaz et~al\mbox{.}(2022)]%
        {fayyaz2022bert}
\bibfield{author}{\bibinfo{person}{Mohsen Fayyaz}, \bibinfo{person}{Ehsan Aghazadeh}, \bibinfo{person}{Ali Modarressi}, \bibinfo{person}{Mohammad~Taher Pilehvar}, \bibinfo{person}{Yadollah Yaghoobzadeh}, {and} \bibinfo{person}{Samira~Ebrahimi Kahou}.} \bibinfo{year}{2022}\natexlab{}.
\newblock \bibinfo{title}{BERT on a Data Diet: Finding Important Examples by Gradient-Based Pruning}.
\newblock
\newblock
\showeprint[arxiv]{2211.05610}~[cs.CL]


\bibitem[Feely et~al\mbox{.}(2023)]%
        {feely-etal-2023-qcon}
\bibfield{author}{\bibinfo{person}{Weston Feely}, \bibinfo{person}{Prabhakar Gupta}, \bibinfo{person}{Manas~Ranjan Mohanty}, \bibinfo{person}{Timothy Chon}, \bibinfo{person}{Tuhin Kundu}, \bibinfo{person}{Vijit Singh}, \bibinfo{person}{Sandeep Atluri}, \bibinfo{person}{Tanya Roosta}, \bibinfo{person}{Viviane Ghaderi}, {and} \bibinfo{person}{Peter Schulam}.} \bibinfo{year}{2023}\natexlab{}.
\newblock \showarticletitle{{QC}on at {S}em{E}val-2023 Task 10: Data Augmentation and Model Ensembling for Detection of Online Sexism}. In \bibinfo{booktitle}{\emph{Proceedings of the 17th International Workshop on Semantic Evaluation (SemEval-2023)}}, \bibfield{editor}{\bibinfo{person}{Atul~Kr. Ojha}, \bibinfo{person}{A.~Seza Do{\u{g}}ru{\"o}z}, \bibinfo{person}{Giovanni Da~San~Martino}, \bibinfo{person}{Harish Tayyar~Madabushi}, \bibinfo{person}{Ritesh Kumar}, {and} \bibinfo{person}{Elisa Sartori}} (Eds.). \bibinfo{publisher}{Association for Computational Linguistics}, \bibinfo{address}{Toronto, Canada}, \bibinfo{pages}{1260--1270}.
\newblock
\urldef\tempurl%
\url{https://doi.org/10.18653/v1/2023.semeval-1.175}
\showDOI{\tempurl}


\bibitem[Guest et~al\mbox{.}(2021)]%
        {guest-etal-2021-expert}
\bibfield{author}{\bibinfo{person}{Ella Guest}, \bibinfo{person}{Bertie Vidgen}, \bibinfo{person}{Alexandros Mittos}, \bibinfo{person}{Nishanth Sastry}, \bibinfo{person}{Gareth Tyson}, {and} \bibinfo{person}{Helen Margetts}.} \bibinfo{year}{2021}\natexlab{}.
\newblock \showarticletitle{An Expert Annotated Dataset for the Detection of Online Misogyny}. In \bibinfo{booktitle}{\emph{Proceedings of the 16th Conference of the European Chapter of the Association for Computational Linguistics: Main Volume}}, \bibfield{editor}{\bibinfo{person}{Paola Merlo}, \bibinfo{person}{Jorg Tiedemann}, {and} \bibinfo{person}{Reut Tsarfaty}} (Eds.). \bibinfo{publisher}{Association for Computational Linguistics}, \bibinfo{address}{Online}, \bibinfo{pages}{1336--1350}.
\newblock
\urldef\tempurl%
\url{https://doi.org/10.18653/v1/2021.eacl-main.114}
\showDOI{\tempurl}


\bibitem[Hampel et~al\mbox{.}(2005)]%
        {bookrobuststatistics}
\bibfield{author}{\bibinfo{person}{Frank Hampel}, \bibinfo{person}{Elvezio Ronchetti}, \bibinfo{person}{Peter Rousseeuw}, {and} \bibinfo{person}{Werner Stahel}.} \bibinfo{year}{2005}\natexlab{}.
\newblock \bibinfo{booktitle}{\emph{Robust Statistics: The Approach Based on Influence Functions}}.
\newblock
\showISBNx{9780471735779}
\urldef\tempurl%
\url{https://doi.org/10.1002/9781118186435}
\showDOI{\tempurl}


\bibitem[Kirk et~al\mbox{.}(2023)]%
        {kirk-etal-2023-semeval}
\bibfield{author}{\bibinfo{person}{Hannah Kirk}, \bibinfo{person}{Wenjie Yin}, \bibinfo{person}{Bertie Vidgen}, {and} \bibinfo{person}{Paul R{\"o}ttger}.} \bibinfo{year}{2023}\natexlab{}.
\newblock \showarticletitle{{S}em{E}val-2023 Task 10: Explainable Detection of Online Sexism}. In \bibinfo{booktitle}{\emph{Proceedings of the 17th International Workshop on Semantic Evaluation (SemEval-2023)}}, \bibfield{editor}{\bibinfo{person}{Atul~Kr. Ojha}, \bibinfo{person}{A.~Seza Do{\u{g}}ru{\"o}z}, \bibinfo{person}{Giovanni Da~San~Martino}, \bibinfo{person}{Harish Tayyar~Madabushi}, \bibinfo{person}{Ritesh Kumar}, {and} \bibinfo{person}{Elisa Sartori}} (Eds.). \bibinfo{publisher}{Association for Computational Linguistics}, \bibinfo{address}{Toronto, Canada}, \bibinfo{pages}{2193--2210}.
\newblock
\urldef\tempurl%
\url{https://doi.org/10.18653/v1/2023.semeval-1.305}
\showDOI{\tempurl}


\bibitem[Koh and Liang(2020)]%
        {koh2020understanding}
\bibfield{author}{\bibinfo{person}{Pang~Wei Koh} {and} \bibinfo{person}{Percy Liang}.} \bibinfo{year}{2020}\natexlab{}.
\newblock \bibinfo{title}{Understanding Black-box Predictions via Influence Functions}.
\newblock
\newblock
\showeprint[arxiv]{1703.04730}~[stat.ML]


\bibitem[Krizhevsky et~al\mbox{.}(2009)]%
        {krizhevskyetal2009}
\bibfield{author}{\bibinfo{person}{Alex Krizhevsky}, \bibinfo{person}{Vinod Nair}, {and} \bibinfo{person}{Geoffrey Hinton}.} \bibinfo{year}{2009}\natexlab{}.
\newblock \showarticletitle{CIFAR-100 (Canadian Institute for Advanced Research)}.
\newblock  (\bibinfo{year}{2009}).
\newblock
\urldef\tempurl%
\url{http://www.cs.toronto.edu/~kriz/cifar.html}
\showURL{%
\tempurl}


\bibitem[Liu et~al\mbox{.}(2019)]%
        {DBLP:journals/corr/abs-1907-11692}
\bibfield{author}{\bibinfo{person}{Yinhan Liu}, \bibinfo{person}{Myle Ott}, \bibinfo{person}{Naman Goyal}, \bibinfo{person}{Jingfei Du}, \bibinfo{person}{Mandar Joshi}, \bibinfo{person}{Danqi Chen}, \bibinfo{person}{Omer Levy}, \bibinfo{person}{Mike Lewis}, \bibinfo{person}{Luke Zettlemoyer}, {and} \bibinfo{person}{Veselin Stoyanov}.} \bibinfo{year}{2019}\natexlab{}.
\newblock \showarticletitle{RoBERTa: {A} Robustly Optimized {BERT} Pretraining Approach}.
\newblock \bibinfo{journal}{\emph{CoRR}}  \bibinfo{volume}{abs/1907.11692} (\bibinfo{year}{2019}).
\newblock
\showeprint[arXiv]{1907.11692}
\urldef\tempurl%
\url{http://arxiv.org/abs/1907.11692}
\showURL{%
\tempurl}


\bibitem[Loshchilov and Hutter(2019)]%
        {loshchilov2019decoupled}
\bibfield{author}{\bibinfo{person}{Ilya Loshchilov} {and} \bibinfo{person}{Frank Hutter}.} \bibinfo{year}{2019}\natexlab{}.
\newblock \bibinfo{title}{Decoupled Weight Decay Regularization}.
\newblock
\newblock
\showeprint[arxiv]{1711.05101}~[cs.LG]


\bibitem[Madukwe et~al\mbox{.}(2020)]%
        {madukwe-etal-2020-data}
\bibfield{author}{\bibinfo{person}{Kosisochukwu Madukwe}, \bibinfo{person}{Xiaoying Gao}, {and} \bibinfo{person}{Bing Xue}.} \bibinfo{year}{2020}\natexlab{}.
\newblock \showarticletitle{In Data We Trust: A Critical Analysis of Hate Speech Detection Datasets}. In \bibinfo{booktitle}{\emph{Proceedings of the Fourth Workshop on Online Abuse and Harms}}, \bibfield{editor}{\bibinfo{person}{Seyi Akiwowo}, \bibinfo{person}{Bertie Vidgen}, \bibinfo{person}{Vinodkumar Prabhakaran}, {and} \bibinfo{person}{Zeerak Waseem}} (Eds.). \bibinfo{publisher}{Association for Computational Linguistics}, \bibinfo{address}{Online}, \bibinfo{pages}{150--161}.
\newblock
\urldef\tempurl%
\url{https://doi.org/10.18653/v1/2020.alw-1.18}
\showDOI{\tempurl}


\bibitem[Mishra and Sachdeva(2020)]%
        {mishra-sachdeva-2020-need}
\bibfield{author}{\bibinfo{person}{Swaroop Mishra} {and} \bibinfo{person}{Bhavdeep~Singh Sachdeva}.} \bibinfo{year}{2020}\natexlab{}.
\newblock \showarticletitle{Do We Need to Create Big Datasets to Learn a Task?}. In \bibinfo{booktitle}{\emph{Proceedings of SustaiNLP: Workshop on Simple and Efficient Natural Language Processing}}, \bibfield{editor}{\bibinfo{person}{Nafise~Sadat Moosavi}, \bibinfo{person}{Angela Fan}, \bibinfo{person}{Vered Shwartz}, \bibinfo{person}{Goran Glava{\v{s}}}, \bibinfo{person}{Shafiq Joty}, \bibinfo{person}{Alex Wang}, {and} \bibinfo{person}{Thomas Wolf}} (Eds.). \bibinfo{publisher}{Association for Computational Linguistics}, \bibinfo{address}{Online}, \bibinfo{pages}{169--173}.
\newblock
\urldef\tempurl%
\url{https://doi.org/10.18653/v1/2020.sustainlp-1.23}
\showDOI{\tempurl}


\bibitem[Muti et~al\mbox{.}(2023)]%
        {muti-etal-2023-uniboes}
\bibfield{author}{\bibinfo{person}{Arianna Muti}, \bibinfo{person}{Francesco Fernicola}, {and} \bibinfo{person}{Alberto Barr{\'o}n-Cede{\~n}o}.} \bibinfo{year}{2023}\natexlab{}.
\newblock \showarticletitle{{U}ni{B}oe{'}s at {S}em{E}val-2023 Task 10: Model-Agnostic Strategies for the Improvement of Hate-Tuned and Generative Models in the Classification of Sexist Posts}. In \bibinfo{booktitle}{\emph{Proceedings of the 17th International Workshop on Semantic Evaluation (SemEval-2023)}}, \bibfield{editor}{\bibinfo{person}{Atul~Kr. Ojha}, \bibinfo{person}{A.~Seza Do{\u{g}}ru{\"o}z}, \bibinfo{person}{Giovanni Da~San~Martino}, \bibinfo{person}{Harish Tayyar~Madabushi}, \bibinfo{person}{Ritesh Kumar}, {and} \bibinfo{person}{Elisa Sartori}} (Eds.). \bibinfo{publisher}{Association for Computational Linguistics}, \bibinfo{address}{Toronto, Canada}, \bibinfo{pages}{1138--1147}.
\newblock
\urldef\tempurl%
\url{https://doi.org/10.18653/v1/2023.semeval-1.158}
\showDOI{\tempurl}


\bibitem[Paul et~al\mbox{.}(2021)]%
        {paul2021deep}
\bibfield{author}{\bibinfo{person}{Mansheej Paul}, \bibinfo{person}{Surya Ganguli}, {and} \bibinfo{person}{Gintare~Karolina Dziugaite}.} \bibinfo{year}{2021}\natexlab{}.
\newblock \bibinfo{title}{Deep Learning on a Data Diet: Finding Important Examples Early in Training}.
\newblock
\newblock
\showeprint[arxiv]{2107.07075}~[cs.LG]


\bibitem[Pruthi et~al\mbox{.}(2020)]%
        {pruthi2020estimating}
\bibfield{author}{\bibinfo{person}{Garima Pruthi}, \bibinfo{person}{Frederick Liu}, \bibinfo{person}{Mukund Sundararajan}, {and} \bibinfo{person}{Satyen Kale}.} \bibinfo{year}{2020}\natexlab{}.
\newblock \bibinfo{title}{Estimating Training Data Influence by Tracing Gradient Descent}.
\newblock
\newblock
\showeprint[arxiv]{2002.08484}~[cs.LG]


\bibitem[Rodríguez-Sánchez et~al\mbox{.}(2022)]%
        {rodriguezsanchezetal}
\bibfield{author}{\bibinfo{person}{Francisco Rodríguez-Sánchez}, \bibinfo{person}{Jorge Carrillo-de Albornoz}, \bibinfo{person}{Laura Plaza}, \bibinfo{person}{Adrián Mendieta-Aragón}, \bibinfo{person}{Guillermo Marco-Remón}, \bibinfo{person}{Maryna Makeienko}, \bibinfo{person}{María Plaza}, \bibinfo{person}{Julio Gonzalo}, \bibinfo{person}{Damiano Spina}, {and} \bibinfo{person}{Paolo Rosso}.} \bibinfo{year}{2022}\natexlab{}.
\newblock \showarticletitle{Overview of EXIST 2022: sEXism Identification in Social neTworks}.
\newblock \bibinfo{journal}{\emph{Procesamiento de Lenguaje Natural}}  \bibinfo{volume}{69} (\bibinfo{date}{09} \bibinfo{year}{2022}), \bibinfo{pages}{229--240}.
\newblock


\bibitem[R{\"o}ttger et~al\mbox{.}(2021)]%
        {rottger-etal-2021-hatecheck}
\bibfield{author}{\bibinfo{person}{Paul R{\"o}ttger}, \bibinfo{person}{Bertie Vidgen}, \bibinfo{person}{Dong Nguyen}, \bibinfo{person}{Zeerak Waseem}, \bibinfo{person}{Helen Margetts}, {and} \bibinfo{person}{Janet Pierrehumbert}.} \bibinfo{year}{2021}\natexlab{}.
\newblock \showarticletitle{{H}ate{C}heck: Functional Tests for Hate Speech Detection Models}. In \bibinfo{booktitle}{\emph{Proceedings of the 59th Annual Meeting of the Association for Computational Linguistics and the 11th International Joint Conference on Natural Language Processing (Volume 1: Long Papers)}}, \bibfield{editor}{\bibinfo{person}{Chengqing Zong}, \bibinfo{person}{Fei Xia}, \bibinfo{person}{Wenjie Li}, {and} \bibinfo{person}{Roberto Navigli}} (Eds.). \bibinfo{publisher}{Association for Computational Linguistics}, \bibinfo{address}{Online}, \bibinfo{pages}{41--58}.
\newblock
\urldef\tempurl%
\url{https://doi.org/10.18653/v1/2021.acl-long.4}
\showDOI{\tempurl}


\bibitem[Samory et~al\mbox{.}(2021)]%
        {samory2021call}
\bibfield{author}{\bibinfo{person}{Mattia Samory}, \bibinfo{person}{Indira Sen}, \bibinfo{person}{Julian Kohne}, \bibinfo{person}{Fabian Floeck}, {and} \bibinfo{person}{Claudia Wagner}.} \bibinfo{year}{2021}\natexlab{}.
\newblock \bibinfo{title}{"Call me sexist, but...": Revisiting Sexism Detection Using Psychological Scales and Adversarial Samples}.
\newblock
\newblock
\showeprint[arxiv]{2004.12764}~[cs.CY]


\bibitem[Sen et~al\mbox{.}(2022)]%
        {sen-etal-2022-counterfactually}
\bibfield{author}{\bibinfo{person}{Indira Sen}, \bibinfo{person}{Mattia Samory}, \bibinfo{person}{Claudia Wagner}, {and} \bibinfo{person}{Isabelle Augenstein}.} \bibinfo{year}{2022}\natexlab{}.
\newblock \showarticletitle{Counterfactually Augmented Data and Unintended Bias: The Case of Sexism and Hate Speech Detection}. In \bibinfo{booktitle}{\emph{Proceedings of the 2022 Conference of the North American Chapter of the Association for Computational Linguistics: Human Language Technologies}}, \bibfield{editor}{\bibinfo{person}{Marine Carpuat}, \bibinfo{person}{Marie-Catherine de~Marneffe}, {and} \bibinfo{person}{Ivan~Vladimir Meza~Ruiz}} (Eds.). \bibinfo{publisher}{Association for Computational Linguistics}, \bibinfo{address}{Seattle, United States}, \bibinfo{pages}{4716--4726}.
\newblock
\urldef\tempurl%
\url{https://doi.org/10.18653/v1/2022.naacl-main.347}
\showDOI{\tempurl}


\bibitem[Simonyan et~al\mbox{.}(2014)]%
        {simonyan2014deep}
\bibfield{author}{\bibinfo{person}{Karen Simonyan}, \bibinfo{person}{Andrea Vedaldi}, {and} \bibinfo{person}{Andrew Zisserman}.} \bibinfo{year}{2014}\natexlab{}.
\newblock \bibinfo{title}{Deep Inside Convolutional Networks: Visualising Image Classification Models and Saliency Maps}.
\newblock
\newblock
\showeprint[arxiv]{1312.6034}~[cs.CV]


\bibitem[Sorscher et~al\mbox{.}(2023)]%
        {sorscher2023neural}
\bibfield{author}{\bibinfo{person}{Ben Sorscher}, \bibinfo{person}{Robert Geirhos}, \bibinfo{person}{Shashank Shekhar}, \bibinfo{person}{Surya Ganguli}, {and} \bibinfo{person}{Ari~S. Morcos}.} \bibinfo{year}{2023}\natexlab{}.
\newblock \bibinfo{title}{Beyond neural scaling laws: beating power law scaling via data pruning}.
\newblock
\newblock
\showeprint[arxiv]{2206.14486}~[cs.LG]


\bibitem[Szegedy et~al\mbox{.}(2015)]%
        {szegedyetal}
\bibfield{author}{\bibinfo{person}{Christian Szegedy}, \bibinfo{person}{Vincent Vanhoucke}, \bibinfo{person}{Sergey Ioffe}, \bibinfo{person}{Jonathon Shlens}, {and} \bibinfo{person}{Zbigniew Wojna}.} \bibinfo{year}{2015}\natexlab{}.
\newblock \showarticletitle{Rethinking the Inception Architecture for Computer Vision}.
\newblock \bibinfo{journal}{\emph{CoRR}}  \bibinfo{volume}{abs/1512.00567} (\bibinfo{year}{2015}).
\newblock
\showeprint[arXiv]{1512.00567}
\urldef\tempurl%
\url{http://arxiv.org/abs/1512.00567}
\showURL{%
\tempurl}


\bibitem[Toneva et~al\mbox{.}(2019)]%
        {toneva2019empirical}
\bibfield{author}{\bibinfo{person}{Mariya Toneva}, \bibinfo{person}{Alessandro Sordoni}, \bibinfo{person}{Remi~Tachet des Combes}, \bibinfo{person}{Adam Trischler}, \bibinfo{person}{Yoshua Bengio}, {and} \bibinfo{person}{Geoffrey~J. Gordon}.} \bibinfo{year}{2019}\natexlab{}.
\newblock \bibinfo{title}{An Empirical Study of Example Forgetting during Deep Neural Network Learning}.
\newblock
\newblock
\showeprint[arxiv]{1812.05159}~[cs.LG]


\bibitem[Vaswani et~al\mbox{.}(2017)]%
        {vaswani2017attention}
\bibfield{author}{\bibinfo{person}{Ashish Vaswani}, \bibinfo{person}{Noam Shazeer}, \bibinfo{person}{Niki Parmar}, \bibinfo{person}{Jakob Uszkoreit}, \bibinfo{person}{Llion Jones}, \bibinfo{person}{Aidan~N. Gomez}, \bibinfo{person}{Lukasz Kaiser}, {and} \bibinfo{person}{Illia Polosukhin}.} \bibinfo{year}{2017}\natexlab{}.
\newblock \bibinfo{title}{Attention Is All You Need}.
\newblock
\newblock
\showeprint[arxiv]{1706.03762}~[cs.CL]


\bibitem[Waseem and Hovy(2016)]%
        {waseem-hovy-2016-hateful}
\bibfield{author}{\bibinfo{person}{Zeerak Waseem} {and} \bibinfo{person}{Dirk Hovy}.} \bibinfo{year}{2016}\natexlab{}.
\newblock \showarticletitle{Hateful Symbols or Hateful People? Predictive Features for Hate Speech Detection on {T}witter}. In \bibinfo{booktitle}{\emph{Proceedings of the {NAACL} Student Research Workshop}}, \bibfield{editor}{\bibinfo{person}{Jacob Andreas}, \bibinfo{person}{Eunsol Choi}, {and} \bibinfo{person}{Angeliki Lazaridou}} (Eds.). \bibinfo{publisher}{Association for Computational Linguistics}, \bibinfo{address}{San Diego, California}, \bibinfo{pages}{88--93}.
\newblock
\urldef\tempurl%
\url{https://doi.org/10.18653/v1/N16-2013}
\showDOI{\tempurl}


\bibitem[Xu et~al\mbox{.}(2020)]%
        {DBLP:journals/corr/abs-2002-10689}
\bibfield{author}{\bibinfo{person}{Yilun Xu}, \bibinfo{person}{Shengjia Zhao}, \bibinfo{person}{Jiaming Song}, \bibinfo{person}{Russell Stewart}, {and} \bibinfo{person}{Stefano Ermon}.} \bibinfo{year}{2020}\natexlab{}.
\newblock \showarticletitle{A Theory of Usable Information Under Computational Constraints}.
\newblock \bibinfo{journal}{\emph{CoRR}}  \bibinfo{volume}{abs/2002.10689} (\bibinfo{year}{2020}).
\newblock
\showeprint[arXiv]{2002.10689}
\urldef\tempurl%
\url{https://arxiv.org/abs/2002.10689}
\showURL{%
\tempurl}


\end{thebibliography}
\clearpage
\appendix
\section{Abbreviations}\label{app: App A.2}
In this Appendix section we will provide the table of abbreviations and their corresponding full forms that have been used in the paper.
\begin{table}[h!]
    \centering
    \footnotesize
    \begin{tabular}{lc}
        \hline
        \textbf{Abbreviation} & \textbf{Full Form}\\ [0.5ex] 
        \hline
        PVI & Pointwise V-Information \\ 
        
        GraNd & Gradient Normed \\
        
        VoG & Variance of Gradients\\
        EL2N & Error L2-Norm \\
        SNLI & Stanford Natural Language Inference \\
        MNLI & Multi-Genre Natural Language Inference \\
        SST-2 & Stanford Sentiment Tree Bank \\
        QNLI & Question-answering Natural Language Inference\\
        EDOS & Explainable Detection of Online Sexism\\
        EXIST & sEXism Identification in Social neTworks \\
        CMSB & Call Me Sexist But\\
        \hline      
    \end{tabular}
    \caption{Table of Abbreviations}
    \label{tab:Table 10}
\end{table}

\section{Dataset Samples}\label{app: App A.1}

In this Appendix section we will provide the top-5 hard and easy examples and their corresponding scores with respect to EL2N and VoG. Tables \ref{tab:table 6} and \ref{tab:table 7} showcase the hard and easy examples based on their respective EL2N scores.
Tables \ref{tab:table 8} and \ref{tab:table 9} contains the hard and easy examples based on their corresponding VoG scores.

\begin{table}[h]
    \footnotesize
    \begin{tabular}{p{6cm} p{0.7cm} p{0.8cm}}
         \textbf{text} & \textbf{gold label} & \textbf{score}\\
         \hline
         MENTION1047 MENTION1497 pls don't the Trump womrn supporters paid or no r so much hotter <hyperlink>
         & sexist & 1.41035\\
         \hline
         SESSIONS IS A FUCKIN PUSSY!! CASE CLOSED!! END OF STORY!! & sexist & 1.40898\\
         \hline
          lmfao i didnt even notice that. I saw she/ and my brain immediately classified that person as retarded and ignored the rest... & sexist & 1.40864\\
         \hline
         Dem FEMBOTS do you really want to protect these guys? ISIS records reveal number of girls raped by each jihadist & sexist & 1.40587 \\
         \hline
         I wonder which of the blonde ones screwed Colin to get that score? Probably both. & sexist & 1.40488\\
         \hline
    \end{tabular}
    \caption{\small{Top 5 high scoring examples (hard) based on EL2N scores.}}
    \label{tab:table 6}
    \end{table}

\begin{table}[b]
    \footnotesize
    \begin{tabular}{p{6cm} p{0.7cm} p{0.8cm}}
         \textbf{text} & \textbf{gold label} & \textbf{score}\\
         \hline
         MENTION3619 MENTION4269 at the very least, it's actionable by twitter. <hyperlink> is good enough for law enforcement.& non-sexist & 0.00185\\
         \hline
         MENTION2715 many people would be scared \& triggered by that cover .. :P & non-sexist & 0.00184\\
         \hline
          MENTION4094 MENTION3619 MENTION378 if you reallllly want that, search for ggautoblocker. i ran some ads last night. & non-sexist & 0.00184\\
         \hline
         MENTION4080 MENTION3619 actually, never mind. no clue who you are. really don't care. go be dramatic in someone else's mentions. & non-sexist & 0.00184 \\
         \hline
          MENTION3273 I guess "no one" includes this entire comment thread: <hyperlink> & non-sexist & 0.00184\\
         \hline
    \end{tabular}
    \caption{\small{Top 5 low scoring examples (easy) based on EL2N scores.}}
    \label{tab:table 7}
    \end{table} 

\begin{table}[b]
    \footnotesize
    \begin{tabular}{p{6cm} p{0.7cm} p{0.8cm}}
         \textbf{text} & \textbf{gold label} & \textbf{score}\\
         \hline
         So....... now........all the rapists and thieves know her name, know her city \& know she is not armed with a weapon........... brilliant......... You know..... left on their own......Liberals would go extinct.....fast & non-sexist & 0.04758\\
         \hline
         Daughter launches vicious knuckleduster attack on woman for 'sleeping with her father'..................... & non-sexist & 0.04528\\
         \hline
          .MENTION1757 MENTION4095 MENTION1563 MENTION2248 MENTION4068 MENTION1148 MENTION3482 Agreed, sure but what does that have to do w \#Page3? & non-sexist & 0.04363\\
         \hline
         But...but...the future is shemale...err...female..! Hillary said... & non-sexist & 0.04219 \\
         \hline
          But don't you miss having a first lady that could beat most men in a bar brawl...or at least B.O...Michael...I mean Michelle...or whatever hi..her name was...was so ma...womanly... that's the one & sexist & 0.04170\\
         \hline
    \end{tabular}
    \caption{\small{Top 5 high scoring examples (hard) based on VoG scores.}}
    \label{tab:table 8}
\end{table}

\begin{table}[b]
    \footnotesize
    \begin{tabular}{p{6cm} p{0.7cm} p{0.8cm}}
         \textbf{text} & \textbf{gold label} & \textbf{score}\\
         \hline
         Danny Devito was separated from his wife because he wouldnâ€™t stop cheating on her with younger women, lol & non-sexist & -0.00445 \\
         \hline
          Hey \#97rock we can't wait until you break out the nigger jokes, Muslim barbs and kike zingers! We await your iconoclastic edginess between repetitive rotations of Faggot Elton John tunes and Born To Be Wild! & non-sexist & -0.00446\\
         \hline
          I ain't gon let these fucken retarded ass mongrels clown me wit dat follow button dog fukdat I'm audi & non-sexist & -0.00451\\
         \hline
        \#mkr I think I just threw up in my mouth :( & non-sexist & -0.00453 \\
         \hline
          MENTION3534 Irony: Dead & non-sexist & -0.00453\\
         \hline
    \end{tabular}
    \caption{\small{Top 5 low scoring examples (easy) based on VoG scores.}}
    \label{tab:table 9}
\end{table}
\end{document}